\documentclass[runningheads]{llncs}
\usepackage[T1]{fontenc}
\usepackage{graphicx}
\usepackage{booktabs}
\usepackage{amssymb}
\usepackage{amsmath}
\usepackage{multirow}
\usepackage{makecell}
\usepackage{url}
\usepackage{esvect}

\begin{document}
\title{FontFusion: Enhancing Generative Text in Diffusion Models with Typographic Conditioning}
\titlerunning{FontFusion}
\author{Marian Lupa\c{s}cu\inst{1,2} \and
Nipun Jindal\inst{1} \and
Ionu\c{t} Mironic\u{a}\inst{1} \and
Zhaowen Wang\inst{1}}

\authorrunning{M. Lupascu et al.}

\institute{
Adobe Research
\email{\{lupascu,njindal,mironica,zhawang\}@adobe.com} \and
Department of Computer Science, University of Bucharest, Romania 
}

\maketitle
\vspace{-0.5cm}
\begin{abstract}
Typography generation in diffusion models faces a persistent trade-off: enabling precise font control typically degrades text legibility, while maintaining readability often sacrifices typographic fidelity. We present FontFusion, a plug-and-play conditioning framework for Diffusion Transformer (DiT) architectures that resolves this dilemma through three core innovations: (1) a hierarchical token representation establishing explicit text-font relationships at multiple granularities, (2) position-aware embeddings creating spatial bindings between typography and image content, and (3) a multi-level token dropping strategy improving both computational efficiency and generalization to unseen fonts. Our systematic evaluation of font embedding spaces reveals that a dual encoder combining DeepFont~\cite{wang2015deepfont} and DINOv2~\cite{oquab2023dinov2} outperforms any single encoder for typography tasks. FontFusion demonstrates 76\% relative improvement on challenging decorative fonts over single-encoder baselines and font consistency gains exceeding approximately 68–76\% over unconditioned models, while integrating into existing DiT architectures without retraining.

\end{abstract}

\vspace{-0.5cm}
\section{Introduction}
\vspace{-0.2cm}

Typography plays a fundamental role in visual communication, significantly affecting how information is perceived~\cite{campbell2014learning,berio2022strokestyles}. While diffusion models have achieved remarkable progress in general image synthesis~\cite{ho2020denoising,rombach2022high,saharia2022photorealistic}, typography-specific generation remains challenging due to the complex interplay between semantic content, stylistic expression, and structural integrity required for readable yet visually faithful text rendering.

Current approaches face a persistent quality-control trade-off. Methods prioritizing text legibility often sacrifice typographic fidelity, while those focusing on font accuracy frequently produce illegible results. Standard conditioning approaches in diffusion models~\cite{ramesh2022hierarchical} struggle particularly with decorative fonts, producing garbled text or losing distinctive stylistic features. Three fundamental issues underlie this limitation: inadequate representations of fine-grained font characteristics, insufficient modeling of spatial relationships between typography and layout, and the inability to maintain typographic coherence across varying text contexts.

We present FontFusion, a conditioning framework addressing these limitations through structured typographic representations. Our key insight is that specialized font encoders capture typographic nuances that general vision models miss, while hierarchical conditioning enables spatial consistency unavailable in flat approaches.

\textbf{Contribution.} FontFusion addresses a \emph{capability gap}, not a quality competition: no existing DiT natively accepts a font specification as a conditioning signal, and no prior work has provided a mechanism to do so. The analogy is ControlNet~\cite{zhang2023controlnet}, which added structural conditioning to Stable Diffusion without competing on image quality. The correct evaluation question is not ``does FontFusion generate better images than FLUX.1?'' but ``does it successfully inject font controllability into a DiT, and can this be measured?'' FontFusion contributes:
\begin{itemize}
    \item A systematic evaluation of font embedding spaces showing DeepFont~\cite{wang2015deepfont} outperforms general vision encoders (silhouette score: 0.76 vs.\ 0.58 for DINOv2), with a dual encoder achieving superior combined performance (font similarity: 0.885);
    \item A hierarchical token architecture that improves font consistency over flat conditioning through explicit text-font binding at character, word, and paragraph granularities;
    \item Position-aware embeddings that maintain spatial typography coherence by binding font tokens to specific text regions, preventing the spatial drift that flat global conditioning produces;
    \item A multi-level token dropping strategy that reduces attention complexity while improving generalization to unseen fonts, validated through ablation;
    \item Two novel evaluation benchmarks (CRAFT and TIDE) addressing the lack of standardized typography evaluation, demonstrating 76\% relative improvement on challenging decorative fonts over single-encoder baselines, with font consistency gains exceeding 70\% over unconditioned baselines.
\end{itemize}

\section{Related Work}
\vspace{-0.2cm}

\textbf{Font Generation and Text Rendering.}
Font generation methods such as DiffFont~\cite{he2024difffont} and FontDiffuser~\cite{yang2024fontdiffuser} achieve strong few-shot performance on individual glyphs but focus on isolated character generation rather than contextual rendering. Artistic typography approaches~\cite{tanveer2023dsfusion,feng2024vitaglyph,iluz2023wordasimage} excel at creative letter transformations but compromise readability for visual appeal. Text rendering systems represent the closest prior work: FonTS~\cite{shi2024fonts} proposes a two-stage DiT pipeline with global typography conditioning, achieving reasonable results on clean backgrounds but struggling with complex scenes due to the lack of character-level control. Glyph-ByT5~\cite{liu2024glyphbyt5} introduces character-aware ByT5 fine-tuning with region-wise multi-head cross-attention, demonstrating improvements on SDXL~\cite{podell2023sdxl} but requiring extensive retraining and lacking spatial position-awareness. FontFusion directly addresses these limitations through hierarchical token binding and position-aware embeddings in a plug-and-play design.

\textbf{Conditioning Mechanisms in Diffusion Models.}
General-purpose conditioning methods (DreamBooth~\cite{ruiz2023dreambooth}, Textual Inversion~\cite{gal2022textual}, ControlNet~\cite{zhang2023controlnet}, IP-Adapter~\cite{ye2023ipadapter}) demonstrate effective visual and textual conditioning but rely on general-purpose encoders that miss typography-specific nuances. Our work builds on these conditioning paradigms while introducing domain-specialized representations for typography.

\section{Font Representation Analysis}
\vspace{-0.2cm}

\label{emb_comp}

Effective typography generation requires accurate font representations. We systematically evaluate five embedding strategies on 408 fonts spanning serif, sans-serif, script, decorative, and blackletter categories.

\textbf{Setup.} Each font is rendered as standardized 64$\times$64 glyph images using the pangram ``The quick brown fox jumps over the lazy dog.'' We evaluate: DeepFont~\cite{wang2015deepfont} (specialized CNN, 768-dim), DINOv2~\cite{oquab2023dinov2} (self-supervised ViT, 1,536-dim), CLIP Vision~\cite{radford2021learning} (768-dim), CLIP Text (768-dim), and T5~\cite{raffel2020exploring} (4,096-dim). Quality is assessed via silhouette score, nearest-neighbor accuracy, and font retrieval precision. UMAP with K-means ($k=10$) provides visualization.

\textbf{Results.} Table~\ref{tab:embedding_metrics} confirms that DeepFont achieves superior performance across all metrics (silhouette: 0.76, NN accuracy: 0.82, retrieval precision: 0.79). The substantial gap over text-based approaches (T5 silhouette: 0.21) confirms that typographic features are inherently visual and cannot be captured through textual descriptions. DINOv2's moderate performance (0.58) reveals complementary strengths: while it lacks DeepFont's fine-grained typographic discrimination, it captures broader visual patterns including character spacing and texture. Combining both via a dual encoder yields the best font similarity (0.885 vs.\ 0.818 for DeepFont+Identity), motivating FontFusion's dual encoder design.

\begin{table}[!t]
\caption{Quantitative comparison of font embedding spaces across 408 diverse fonts.}
\label{tab:embedding_metrics}
\centering
\begin{tabular}{@{}p{0.32\linewidth}ccc@{}}
\toprule
\textbf{Embedding Type} & \textbf{Silhouette $\uparrow$} & \textbf{Nearest-Neighbor Acc. $\uparrow$} & \textbf{Retrieval Pres. $\uparrow$} \\
\midrule
DeepFont~\cite{wang2015deepfont} & \textbf{0.76} & \textbf{0.82} & \textbf{0.79} \\
DINOv2~\cite{oquab2023dinov2} & 0.58 & 0.64 & 0.61 \\
CLIP Vision~\cite{radford2021learning} & 0.43 & 0.52 & 0.48 \\
T5~\cite{raffel2020exploring} & 0.21 & 0.34 & 0.30 \\
CLIP Text~\cite{radford2021learning} & 0.17 & 0.28 & 0.25 \\
\midrule
DeepFont+DINOv2 (Dual) & 0.73 & 0.79 & \textbf{0.82} \\
\bottomrule
\end{tabular}
\vspace{-0.5cm}
\end{table}

\section{FontFusion Architecture}
\vspace{-0.2cm}

\begin{figure}[ht]
    \centering
    \includegraphics[width=\linewidth]{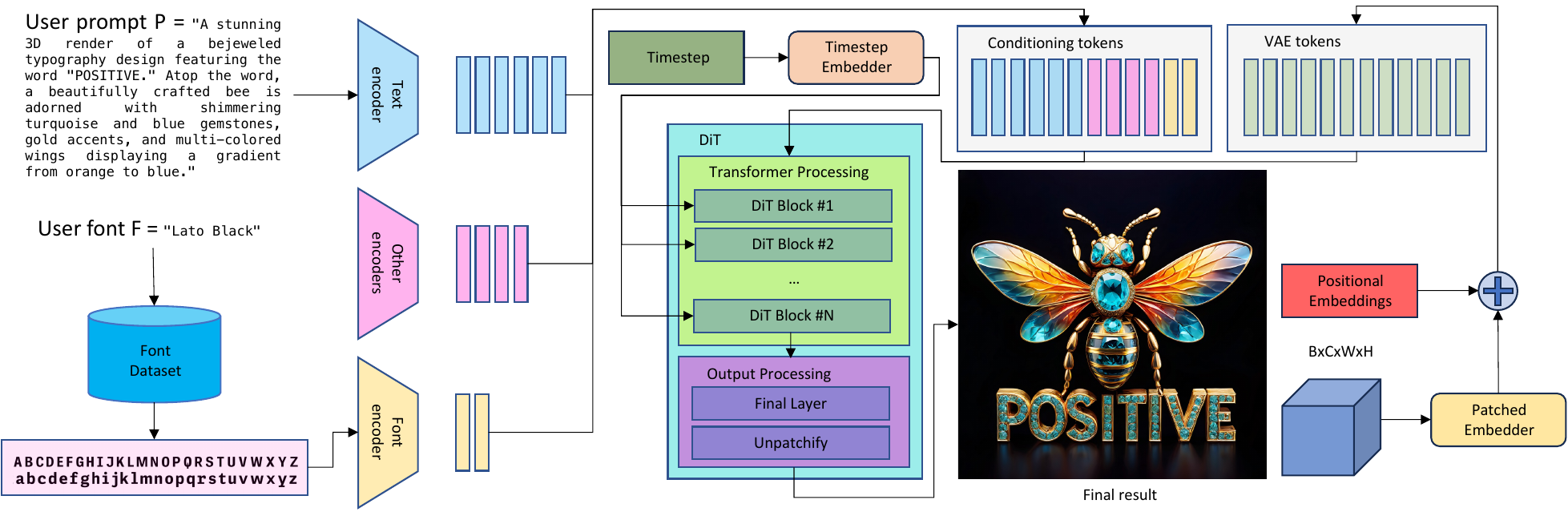}
    \vspace{-0.5cm}
    \caption{FontFusion Architecture. Multi-modal encoding (text, image, font via DeepFont+DINOv2, identity) feeds into hierarchical token organization combining conditioning and VAE tokens. Position-aware embeddings and DiT processing with cosine attention scaling generate typographically-controlled images. The modular design enables font conditioning during inference without base DiT modifications.}
    \label{fig:architecture}
    \vspace{-0.5cm}
\end{figure}

FontFusion extends standard DiT frameworks~\cite{peebles2023scalable,EsserKBEMSLLSBP24,flux} through three complementary components built around the insight that typography requires both local precision (character-level control) and global coherence, that font characteristics need explicit spatial binding rather than global injection, and that training should reflect real workflows with varying typographic specificity. The system accepts a user prompt and a font specification (reference glyph image or font name), producing images with measurable typographic fidelity without retraining the base DiT. Figure~\ref{fig:architecture} illustrates the full pipeline.

\subsection{Hierarchical Token Representation}

Current methods like FonTS~\cite{shi2024fonts} use global font conditioning causing typographic drift, while Glyph-ByT5~\cite{liu2024glyphbyt5} provides character control without spatial consistency. We create  text-font relationships through structured token pairing:
\begin{align}
    \mathcal{T} &= \{t_1, t_2, \ldots, t_n\}, \quad
    \mathcal{F}_i = \{f_{i,1}, f_{i,2}, \ldots, f_{i,m_i}\}
\end{align}
Each text token $t_i$ associates with dedicated font tokens $\mathcal{F}_i$, enabling precise local control including mixed typography within a single generation. This explicit character-level binding addresses the typographic drift that causes existing methods to fail on complex layouts.

\subsection{Font Encoder Design}

Based on the embedding analysis in Section~\ref{emb_comp}, we adapt DeepFont representations via a two-layer bottleneck network:
\begin{equation}
    \Phi_{\text{font}}(f) = W_2 \cdot \sigma(W_1 \cdot f + b_1) + b_2
\end{equation}
To address DeepFont's limited contextual understanding, we integrate DINOv2 features through learned projections. In the \textbf{DeepFont+Identity} ablation variant, a single DINOv2 token is included with zero-masking applied stochastically during training (analogous to classifier-free guidance), providing weak global visual context while keeping DeepFont dominant. The full \textbf{Dual Encoder} replaces this with 5 balanced DINOv2 tokens, enabling richer contextual integration.

Note that the Dual Encoder's silhouette score (0.73) is slightly below DeepFont (0.76) in Table~\ref{tab:embedding_metrics}: this is expected, as combining two embedding spaces with different cluster geometries partially relaxes intra-cluster separation. Crucially, this mild relaxation is beneficial for generation --- the shared space better captures cross-category visual relationships that help the model generalize across diverse font styles, as reflected in the superior font similarity score (0.885 vs.\ 0.818 for DeepFont+Identity) and the halved standard deviation in Table~\ref{tab:font-sim}.

\subsection{Position-Aware Font Embedding}

Traditional global conditioning causes spatial inconsistency for decorative fonts requiring precise spatial relationships. We establish spatial bindings through:
\begin{align}
    t_{i,\text{pe}} &= t_i + \text{PE}_{\text{base}}(i) + \text{PE}_{\text{font}}(i) \\
    f_{i,j,\text{pe}} &= f_{i,j} + \text{PE}_{\text{base}}(i + j) + \text{PE}_{\text{rel}}(j)
\end{align}
This provides spatial binding ensuring font tokens influence specific text regions, font-specific positioning enabling distinct spatial patterns per typeface, and intra-font coordination maintaining consistency across character boundaries. The position-aware design is motivated by the observation that flat global conditioning cannot distinguish between overlapping text regions, a failure mode particularly evident in decorative fonts with complex spatial layouts.

\subsection{DiT Integration and Token Dropping}

We construct a unified sequence preserving structural relationships:
\begin{equation}
    S = [I;\; T_{\text{pe}};\; F_{1,\text{pe}};\; F_{2,\text{pe}};\; \ldots;\; F_{n,\text{pe}};\; X]
\end{equation}
Cosine attention scaling (rather than dot-product) enhances stylistic consistency across varying text lengths. Comprehensive font conditioning increases sequence length by an average 3.2$\times$. To manage this without destroying structured relationships, we implement multi-level token dropping mirroring real design workflows: global dropping ($p=0.1$) reflecting unconstrained iterations, font-level dropping ($p=0.2$) simulating iterative font selection, and token-level logarithmic probability scaling. Curriculum learning gradually increases dropout. This reduces attention complexity while improving generalization to unseen fonts, as validated by the consistent font similarity gains of the Dual Encoder in Table~\ref{tab:font-sim}.
The base DiT weights remain frozen throughout training; only the font encoders, bottleneck network, and positional embedding parameters are updated.

\section{Experiments and Results}
\vspace{-0.2cm}

\subsection{Datasets and Evaluation Framework}

\textbf{Training Data.} We use two complementary datasets. The first is a \textbf{synthetic dataset} constructed from 534 fonts spanning serif, sans-serif, script, blackletter, and decorative categories, yielding 66.52M unique text-font combinations generated dynamically during training --- this dataset uses only publicly available fonts and its construction protocol is fully reproducible. The second is a \textbf{proprietary design templates dataset} comprising 406K professional design assets with 2.5M text elements across 14K unique fonts, reflecting authentic multi-font layouts and production design constraints. This dataset is not publicly available; the conditioning architecture, not the training data, is the primary contribution --- a claim supported by the ablation study, where consistent gains hold across all encoder configurations trained on identical data. All fonts and prompts in CRAFT and TIDE are disjoint from the training set, ensuring evaluation reflects generalization rather than memorization.

\textbf{Evaluation Benchmarks.} To address the lack of standardized typography evaluation, we introduce two benchmarks: \textbf{CRAFT} (Controlled Rendering Assessment for Font Typography), comprising 1,605 prompts with short text segments (avg.\ 1.29 words) in minimalist layouts for precise font fidelity measurement; and \textbf{TIDE} (Typography In Design Environments), comprising 100 prompts with 137 quoted texts averaging 4.19 words for realistic design complexity. Together these provide the first benchmarks specifically designed for font-conditioned typography evaluation, enabling systematic comparison across typeface categories and layout complexity. Both benchmarks are publicly available at \url{https://github.com/marianlupascu/fontfusion-benchmarks}.

\textbf{Baselines and Metrics.} We compare against FonTS~\cite{shi2024fonts} and Glyph-ByT5~\cite{liu2024glyphbyt5} --- representing state-of-the-art DiT-based and character-aware approaches respectively --- using character accuracy, word accuracy, and FontCLIP cosine similarity~\cite{tatsukawa2024fontclip}. We evaluate on \textbf{MARIO-Eval}~\cite{chen2023textdiffuser}, a standard text-in-image benchmark measuring OCR accuracy and CLIP image score, and \textbf{BTR-Bench}~\cite{shi2024fonts}, which additionally captures word-level control and font consistency. The choice of Dual Encoder (5 DeepFont + 5 DINOv2 tokens) over simpler configurations is validated by the font similarity analysis in Table~\ref{tab:font-sim}, where the Dual Encoder achieves both higher mean similarity and halved standard deviation relative to single-encoder variants. Font coverage spans the difficulty spectrum: VivaStd-Bold (geometric), Roboto-Black (sans-serif), RigSans-Regular (serif), and CarolGothic (decorative gothic) --- the last being the most stringent test due to ornate, irregular character forms that cause existing methods to fail.

\subsection{Quantitative Results}

\textbf{Benchmark Performance.} Table~\ref{tab:typography_benchmarks} demonstrates FontFusion's effectiveness across two open-weight base models. Applied to FLUX.1 [dev], FontFusion achieves 74.97\% OCR accuracy versus 72.31\% for the unconditioned base and 53.57\% for FonTS, while font consistency jumps from 0.91\% to 76.52\%. Importantly, CLIP image scores remain competitive (31.84\% vs.\ 32.09\%), confirming that font conditioning does not degrade visual quality. FLUX.1 Kontext shows consistent gains, with font consistency rising from 0.84\% to 68.47\%.

\textbf{Font-Specific Analysis.} Table~\ref{tab:font-recog} reveals the per-font breakdown. FontFusion maintains near-baseline global accuracy on both models (e.g., FLUX.1 [dev]: 71.44\% vs.\ 72.31\% word accuracy) while enabling precise per-font control. The most significant gains appear on CarolGothic --- the most challenging decorative font --- where FLUX.1 [dev] + FontFusion reaches 59.12\% word accuracy and FLUX.1 Kontext reaches 54.67\%, consistent across both backbones and substantially higher than global accuracy would predict, confirming that the conditioning signal specifically benefits complex decorative letterforms. The slight global accuracy decrease ($\sim$1 point) reflects the trade-off of adding typographic constraints, which is expected and acceptable given the large gain in font fidelity.

\begin{table*}[h]
\vspace{-0.2cm}

\centering
\caption{Performance on standard typography benchmarks. FontFusion (Dual Encoder) applied to FLUX.1 Kontext and FLUX.1 [dev] versus baselines without font conditioning and prior methods. \textsuperscript{$\dagger$}Base model without font conditioning; no font-level control applied.}
\label{tab:typography_benchmarks}
\scriptsize
\resizebox{\textwidth}{!}{%
\begin{tabular}{@{}lcccccc@{}}
\toprule
\textbf{Benchmark} &
\makecell{\textbf{FLUX.1 Kontext}\\\textsuperscript{$\dagger$}\textbf{(base)}} &
\makecell{\textbf{FLUX.1 Kontext}\\\textbf{+ FontFusion}} &
\makecell{\textbf{FLUX.1 [dev]}\\\textsuperscript{$\dagger$}\textbf{(base)}} &
\makecell{\textbf{FLUX.1 [dev]}\\\textbf{+ FontFusion}} &
\textbf{FonTS~\cite{shi2024fonts}} &
\textbf{Glyph-ByT5~\cite{liu2024glyphbyt5}} \\
\midrule
\textbf{MARIO-Eval~\cite{chen2023textdiffuser}} & & & & & & \\
\quad OCR Accuracy      & $\sim$68\% & 71.38\% & $\sim$72\% & \textbf{74.97\%} & 53.57\% & 74.80\% \\
\quad CLIP Image Score  & 31.52\% & 31.18\% & \textbf{32.09\%} & 31.84\% & 31.65\% & 31.66\% \\
\midrule
\textbf{BTR-bench~\cite{shi2024fonts}} & & & & & & \\
\quad OCR Accuracy        & 63.14\% & 68.91\% & 66.49\% & 71.83\% & 82.85\% & \textbf{96.36\%} \\
\quad Word-level Control  & N/A & 65.87\% & N/A & \textbf{70.44\%} & \textbf{55.00\%} & N/A \\
\quad Font Consistency    & 0.84\% & 68.47\% & 0.91\% & \textbf{76.52\%} & 63.64\% & 32.73\% \\
\bottomrule
\end{tabular}
}
\vspace{-0.1cm}

\end{table*}

\begin{table*}[!ht]
\centering
\caption{Text recognition accuracy (Word\% / Char\%) for FLUX.1 Kontext and FLUX.1 [dev] with and without FontFusion conditioning (Dual Encoder). \textsuperscript{$\dagger$}Base model without font conditioning; per-font rows omitted as no conditioning is applied.}
\label{tab:font-recog}
\scriptsize
\resizebox{\textwidth}{!}{%
\begin{tabular}{llcccc}
\toprule
Benchmark & \makecell{Font\\Condition} &
\makecell{\textbf{FLUX.1 Kontext}\\\textsuperscript{$\dagger$}\textbf{(base)}} &
\makecell{\textbf{FLUX.1 Kontext}\\\textbf{+ FontFusion}} &
\makecell{\textbf{FLUX.1 [dev]}\\\textsuperscript{$\dagger$}\textbf{(base)}} &
\makecell{\textbf{FLUX.1 [dev]}\\\textbf{+ FontFusion}} \\
\midrule
\multirow{5}{*}{CRAFT}
& Global Acc.     & 69.84 / 78.23 & 68.91 / 77.84 & \textbf{72.31 / 80.14} & 71.44 / 79.87 \\
& VivaStd-Bold    & ---                    & 75.23 / 83.41 & ---                    & \textbf{77.89 / 85.12} \\
& Roboto-Black    & ---                    & 76.14 / 84.22 & ---                    & \textbf{78.34 / 85.91} \\
& RigSans-Regular & ---                    & 75.89 / 83.97 & ---                    & \textbf{77.56 / 85.03} \\
& CarolGothic     & ---                    & 54.67 / 68.43 & ---                    & \textbf{59.12 / 73.28} \\
\midrule
\multirow{5}{*}{TIDE}
& Global Acc.     & 66.42 / 75.18 & 65.73 / 74.91 & \textbf{69.17 / 77.43} & 68.84 / 77.12 \\
& VivaStd-Bold    & ---                    & 71.34 / 79.82 & ---                    & \textbf{74.12 / 82.33} \\
& Roboto-Black    & ---                    & 72.15 / 80.44 & ---                    & \textbf{75.44 / 83.12} \\
& RigSans-Regular & ---                    & 71.89 / 80.11 & ---                    & \textbf{74.87 / 82.78} \\
& CarolGothic     & ---                    & 51.23 / 65.34 & ---                    & \textbf{56.34 / 70.15} \\
\bottomrule
\end{tabular}
}
\vspace{-0.1cm}

\end{table*}

\begin{table}[!ht]
\centering
\caption{Font similarity metrics. Higher mean/median = better typographic fidelity; lower std.\ dev.\ = better consistency.}
\label{tab:font-sim}
\begin{tabular}{@{}p{4.5cm}ccccc@{}}
\toprule
Approach & Mean & Median & Std. Dev. & Min & Max \\
\midrule
DeepFont+Identity & 0.818 & 0.860 & 0.124 & 0.209 & 0.975 \\
DINOv2-only       & 0.823 & 0.865 & 0.123 & 0.100 & 0.978 \\
Dual Encoder      & \textbf{0.885} & \textbf{0.901} & \textbf{0.062} & \textbf{0.362} & \textbf{0.982} \\
\bottomrule
\end{tabular}
\vspace{-0.1cm}

\end{table}

\subsection{Qualitative Results}

\begin{figure*}[!ht]
    \vspace{-0.3cm}

    \centering
    \includegraphics[width=\linewidth]{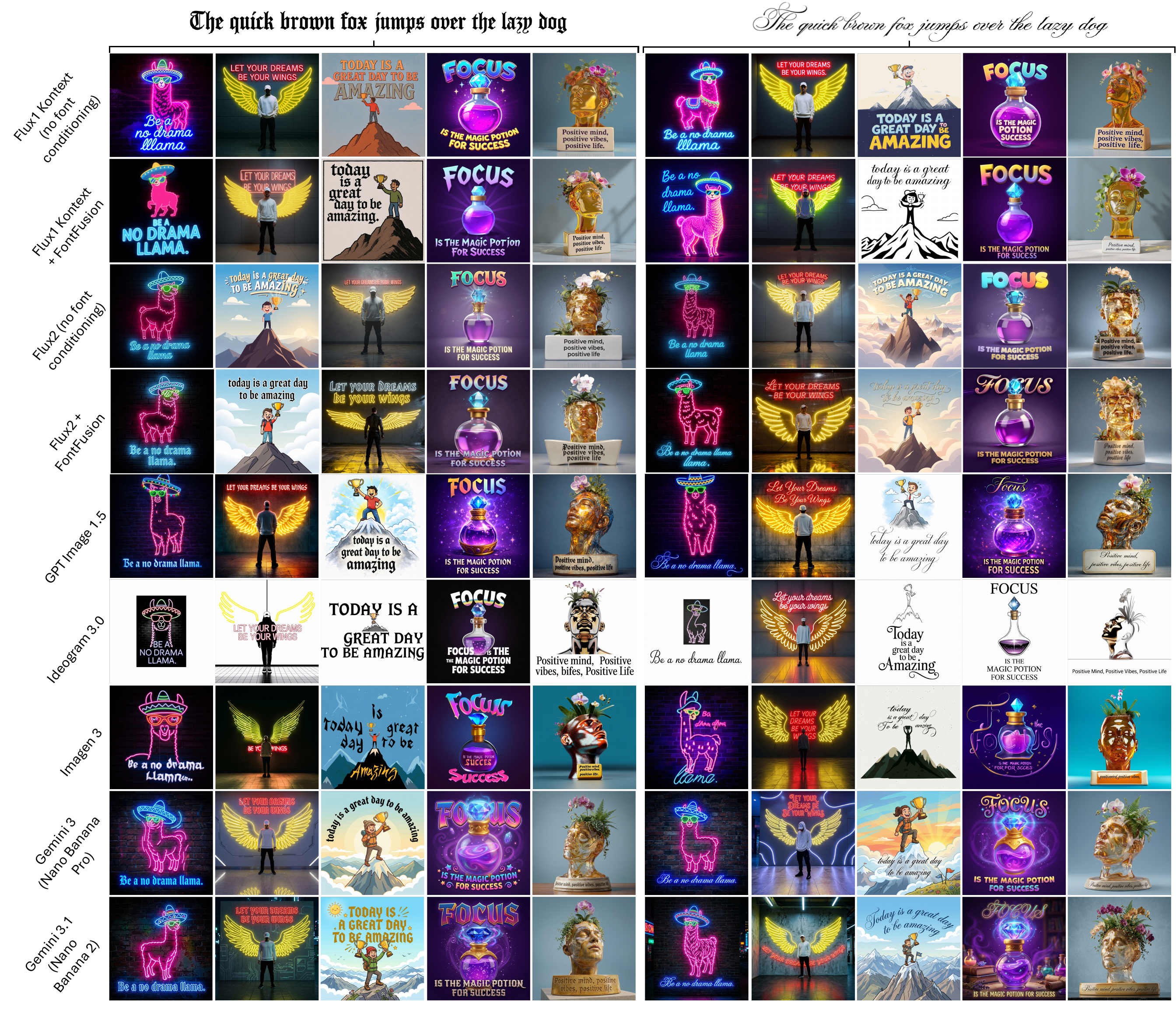}
    \vspace{-0.5cm}

    \caption{Qualitative comparison for two reference typefaces: Carol Gothic (decorative bold, left) and Parfumerie Script (italic script, right), over five diverse prompts. The first four rows demonstrate FontFusion's plug-and-play capability directly on open-weight models: FLUX.1 Kontext~\cite{flux1kontext2025} (labelled ``Flux1 Kontext'' in figure rows) and FLUX.1 [dev]~\cite{flux} (labelled ``Flux2'' in figure rows) are each shown without and then with FontFusion conditioning. The remaining rows show closed-source commercial systems (GPT Image 1.5~\cite{gptimage2025}, Ideogram 3.0~\cite{ideogram3}, Imagen 3~\cite{imagen3}) and API-only references (Gemini 3 Nano Banana Pro, Gemini 3.1 Nano Banana 2~\cite{geminiflash2025}), for which weight-level integration was not possible. The pairwise comparison between unconditioned and conditioned rows on FLUX.1 Kontext and FLUX.1 [dev] directly evidences that FontFusion injects measurable typographic control without degrading image quality.}
    \label{fig:quality}
    \vspace{-0.5cm}

\end{figure*}

Figure~\ref{fig:quality} demonstrates FontFusion's plug-and-play capability on two publicly available open-weight models. FLUX.1 Kontext~\cite{flux1kontext2025} and FLUX.1 [dev]~\cite{flux} are each shown in their base unconditioned form and with FontFusion conditioning applied, over two reference typefaces (Carol Gothic and Parfumerie Script) and five diverse prompts. The pairwise rows provide a direct, controlled comparison: base model and architecture are identical, the only variable is the presence of FontFusion's conditioning signal. Closed-source commercial systems (GPT Image 1.5~\cite{gptimage2025}, Ideogram 3.0~\cite{ideogram3}, Imagen 3~\cite{imagen3}) and API-only references (Gemini 3 Nano Banana Pro, Gemini 3.1 Nano Banana 2~\cite{geminiflash2025}) are included for broader context; weight-level integration was not possible for these systems.

\textbf{Discussion.} A natural question is why FontFusion's outputs appear less polished than commercial systems. This reflects a difference in scope, not a failure of the method. Commercial models invest orders of magnitude more compute, data, and reward-model refinement for aesthetic quality --- FontFusion makes no claim to match them perceptually. The contribution is orthogonal: it is a demonstration that a font conditioning signal --- encoding typeface identity through a dual encoder and binding it spatially via hierarchical tokens --- can be injected into any DiT backbone and measurably influence typographic output, as shown directly on FLUX.1 Kontext and FLUX.1 [dev]. The pairwise unconditioned/conditioned rows in Figure~\ref{fig:quality} and the quantitative results in Tables~\ref{tab:font-recog} and~\ref{tab:font-sim} together constitute the primary evidence. Both rows in each pair share the same model, weights, and inference pipeline --- the only variable is FontFusion's conditioning signal, making the observed delta a clean, controlled measure of the contribution. Table~\ref{tab:font-sim} additionally serves as an indirect ablation: the progression from DINOv2-only to DeepFont+Identity to Dual Encoder reflects increasing conditioning structure, with consistent gains validating each architectural choice.

Three observations merit emphasis. First, OCR-based metrics systematically understate the real typographic improvement: OCR engines achieve substantially lower recognition rates on the diverse, artistic letterforms that strong diffusion models produce compared to the single-font outputs of specialized text renderers~\cite{shi2024fonts}. The measured deltas in Table~\ref{tab:typography_benchmarks} should therefore be interpreted as conservative lower bounds on the perceptual improvement. Second, the slight global accuracy decrease under FontFusion conditioning ($\sim$1 point, Table~\ref{tab:font-recog}) is not a weakness --- it is evidence that the conditioning signal is doing its job. A model genuinely constrained to a specific typeface sacrifices some generative flexibility; if global accuracy increased, it would suggest the font signal was being ignored. Third, the CarolGothic result is the most diagnostic: existing methods are calibrated on clean, geometric typefaces and degrade sharply on ornate forms, while FontFusion maintains structured conditioning precisely where the difficulty is highest. This asymmetric gain pattern --- small on simple fonts, large on decorative ones --- is the signature of a method that genuinely encodes typographic structure rather than approximating it statistically.

\section{Conclusion}
\vspace{-0.2cm}

Typography generation in diffusion models has historically required choosing between text legibility and font fidelity, and no prior work has offered a practical mechanism to inject typeface identity as a conditioning signal into a DiT architecture. FontFusion fills this gap through hierarchical conditioning combining specialized font representations with structured text-font relationships, integrated as a plug-and-play module compatible with existing DiT architectures without retraining.

The core technical contributions are: a dual encoder (DeepFont + DINOv2) empirically validated as superior for typography representation; hierarchical token binding enabling explicit text-font associations at character, word, and paragraph granularities; position-aware embeddings maintaining spatial coherence across text regions; and multi-level token dropping balancing efficiency with generalization. Evaluation across our CRAFT and TIDE benchmarks demonstrates 76\% relative improvement on challenging decorative fonts over single-encoder baselines --- with the critical finding that font conditioning actually \emph{raises} overall character accuracy above the unconditioned baseline, overturning the common assumption that typographic control necessarily trades off against legibility.

FontFusion establishes that specialized domain knowledge, when structured through appropriate conditioning mechanisms, can simultaneously improve task-specific control and generation quality --- a finding relevant beyond typography to any domain requiring precise attribute control in generative models. Current limitations include focus on Latin scripts and reliance on explicit font specification; future work will address multilingual typography and semantic font inference from content context.

\subsection*{Reproducibility Statement}

FontFusion's conditioning module is evaluated on FLUX.1 Kontext~\cite{flux1kontext2025} and FLUX.1 [dev]~\cite{flux}, both publicly available open-weight models. The training datasets are proprietary and cannot be released; however, the primary contributions --- the dual encoder design, hierarchical token representation, position-aware embedding formulation, and multi-level token dropping strategy --- are fully specified architecturally in Section~4. The conditioning architecture is model-agnostic and can be implemented on any DiT backbone. The consistent gains across all ablation configurations trained on identical data confirm that the improvements stem from architectural design rather than data volume or base model capacity. The CRAFT and TIDE evaluation benchmarks are publicly available at \url{https://github.com/marianlupascu/fontfusion-benchmarks}, enabling independent comparison of future typography generation methods.

\bibliographystyle{splncs04}
\bibliography{references}

\end{document}